%% file: main.tex
\documentclass[letterpaper, 10 pt, conference]{ieeeconf}  

\IEEEoverridecommandlockouts                              
\input{preamble.tex}
\title{\LARGE \bf 
    Radar Odometry Subject to High Tilt Dynamics of Subarctic Environments
}

\author{
    Matěj Boxan,$^{1}$ William Larrivée-Hardy,$^{1}$ and François Pomerleau$^{1}$
    \thanks{
        \raggedright$^{1}$ Northern Robotics Laboratory, Université Laval, Quebec, Canada.
    }
    \thanks{
        {\texttt{\small{matej.boxan@norlab.ulaval.ca, francois.pomerleau@ift.ulaval.ca}}}    
    }
}

\begin{document}

\maketitle
\thispagestyle{empty}
\pagestyle{empty}

\begin{abstract}
Rotating FMCW radar odometry methods often assume flat ground conditions.
While this assumption is sufficient in many scenarios, including urban environments or flat mining setups, the highly dynamic terrain of subarctic environments poses a challenge to standard feature extraction and state estimation techniques.
This paper benchmarks three existing radar odometry methods under demanding conditions, exhibiting up to \qty{13}{\degree} in pitch and \qty{4}{\degree} in roll difference between consecutive scans, with absolute pitch and roll reaching \qty{30}{\degree} and \qty{8}{\degree}, respectively.
Furthermore, we propose a novel radar-inertial odometry method utilizing tilt-proximity submap search and a hard threshold for vertical displacement between scan points and the estimated axis of rotation.
Experimental results demonstrate a state-of-the-art performance of our method on an urban baseline and a 0.3\% improvement over the second-best comparative method on a 2-kilometer-long dynamic trajectory.
Finally, we analyze the performance of the four evaluated methods on a complex radar sequence characterized by high lateral slip and a steep ditch traversal.
\end{abstract}

\section{Introduction}

Recent years have seen an increase deployment of radar technologies in various conditions \cite{Harlow2024_newwave}.
Compared to lidars and cameras, radar sensors are immune to illumination changes, atmospheric conditions and small particles in the air.
This robustness to external factors make radars well suited for navigation in challenging conditions, such as in the mining industry.
However, radar sensors face numerous challenges, including low signal-to-noise ratio, low data rates, and the 2D nature of today's \ac{FMCW} radars.

\begin{figure}
    \centering
    \includegraphics[width=\linewidth]{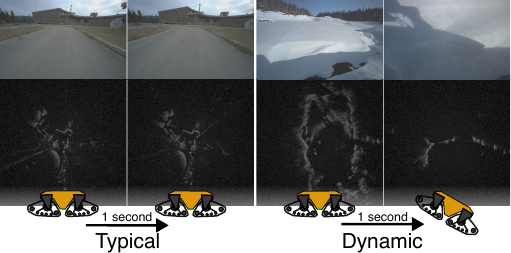}
    \caption{Typical rotating \acs{FMCW} radar solutions operate in flat terrains, where the environments and the vehicle's mobility do not allow sudden changes in the vehicles tilt.
    In contrast, we evaluate radar odometry methods in highly dynamic environments, where the difference in pitch in the span of \qty{1}{\second} can be as high as \qty{30}{\degree}.
    The top row provides a context for the Cartesian radar scan from a front mounted camera.
    }
    \label{fig:intro}
\end{figure}

Multiple radar odometry estimation methods were proposed in this context.
Through efficient feature extraction~\cite{Adolfsson2023_cfear}, ORB descriptor matching in the radar image space \cite{Lim2023_orora} or an adaptive voting strategy \cite{Yang2025_rino}, researchers were able to achieve lidar-level performance with rotating~\ac{FMCW} radar sensor.
However, current solutions evaluate their state estimation performance on urban sequences that include flat ground and distinct features, such as the Boreas~\cite{Burnett2023_boreas} or Oxford RobotCar~\cite{Maddern2017} datasets.
While these datasets provide a great benchmarking infrastructure for odometry performance comparison, they do not necessarily verify the algorithms’ robustness to dynamic events, such as sudden vehicle tilt changes, as shown in \autoref{fig:intro}.

In this paper, we benchmark three existing radar-odometry methods on demanding sequences from the \ac{FoMo} dataset~\cite{Boxan2026_fomo}.
These sequences exhibit high pitch and roll variations, including differences of up to \qty{13}{\degree} in pitch and \qty{4}{\degree} in roll between consecutive scans, with absolute pitch and roll reaching \qty{30}{\degree} and \qty{8}{\degree}, respectively.
Furthermore, we propose a novel radar-inertial odometry method utilizing tilt-proximity submap search and a hard threshold for vertical displacement between scan points and the estimated axis of rotation.
In summary, the contributions of our work are:

\begin{itemize}
    \item A radar-inertial odometry method employing tilt-proximity search for \acs{ICP} reference maps;
    \item A hard threshold filter based on the Cauchy function of the vertical displacement between radar scan points and an estimated axis of rotation;
    \item An evaluation of our method alongside three existing state-of-the-art methods on challenging dynamic sequences from the \ac{FoMo} dataset.
\end{itemize}

\section{Related Work}

\begin{figure*}[t]
    \centering
    \resizebox{\textwidth}{!}{\includestandalone{diagram}}
    \\[3pt]
    \includegraphics[width=\textwidth]{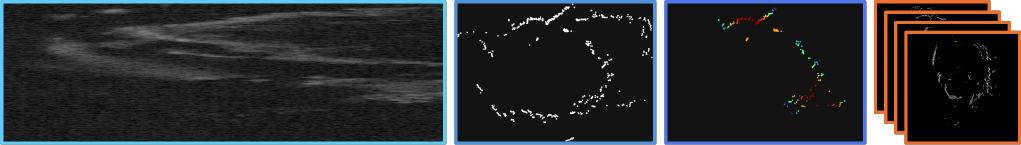}
    \caption{A diagram of our method's pipeline.
    Bottom images show different stages of the processed radar data, where the border colors match the corresponding arrows in the diagram.
    The images are cropped to information-rich sections.
    }
    \label{fig:method}
\end{figure*}

Existing research in radar odometry can be broadly categorized into two main clusters: dense and sparse methods.
Dense methods, such as the work of \citet{Park2020_pharao}, operate on full radar images, attempting to find the relative motion between consecutive scans that maximizes their cross-correlation.
They leverage the full environmental context encoded in the raw radar data, but can be computationally expensive.
In contrast, sparse methods reduce the continuous power spectrum of the raw radar data into a set of geometric features, before passing them to a matching backend, such as the \ac{ICP}.
The performance of sparse techniques is inherently tied to the feature extractor.
Popular extractors include the \ac{CFAR}-based approach utilized by \citet{Vivet2013_localization} and the \textit{k}-strongest method by \citet{Adolfsson2023_cfear}.
Because radar data are notoriously prone to artifacts and multi-path reflections, robust outlier rejection is critical.
For instance, ORORA \cite{Lim2023_orora} improves registration by decoupling rotation and translation while accounting for the anisotropic nature of radar noise. This was further evolved into RINO \cite{Yang2025_rino}, which introduces an adaptive voting mechanism and motion compensation within a loosely coupled radar-inertial framework.
Despite these advancements, many methods struggle with accurate heading estimation, particularly in unstructured environments.
While \ac{RTR} \cite{Qiao2025} attempts to mitigate this by using the \ac{IMU}'s vertical-axis angular velocity as an orientation prior, this single-axis assumption fails in complex terrain where significant pitch and roll variations are present.
Such pronounced vehicle attitudes not only degrade heading priors but also alter the observed geometry between scans, leading to data association failures.
To address these challenges, we propose a radar-inertial odometry method that utilizes a tilt-compensated submap search and an enhanced scan-filtering strategy to maintain robustness in high-complexity, dynamic environments.

\section{Methodology}

Our method combines a classic feature extraction approach and a Point-to-point \ac{ICP} with tilt-based submap search and point filtering.
The complete pipeline is presented in \autoref{fig:method}.
In green, the diagram shows processing of the \ac{IMU} measurements, provided at \qty{200}{\hertz}. 
The IMU data are unbiased using an average over a \qty{10}{\second}-long initial static window.
Afterwards, the linear accelerations $\mathbf{a}$ and angular velocities $\mathbf{\omega}$ are fed into a Madgwick filter \cite{Madgwick2011} to estimate the vehicle's orientation in~3D as $q\in\mathbb{H}$.
We use the orientation corresponding to the start of the radar scan to perform a correspondence search in an array of past submap.
The spatially closest submap withing a distance window $r_\text{submap}$ and roll and pitch difference under the tilt threshold $\tau_\text{tilt}$ is kept.
This search provides the relative orientation $q_r$, together with a submap point cloud $\mathcal{Q}$.
If no matching submap is found, we employ a conservative approach, resetting the estimated vehicle velocity to $\mathbf{0}$.

The radar data processing is shown in blue.
We start with raw \qty{360}{\degree} polar intensity images $Z_{N_a\times N_r}$ with $N_a$ azimuth and $N_r$ range bins, provided at \qty{4}{\hertz}.
Features are extracted with the \textit{k}-strongest approach, as proposed by \citet{Adolfsson2023_cfear}.
We only keep return between $r_\text{min}$ and $r_\text{max}$ distance from the sensor, and extract the $k$ high intensity peaks per azimuth with higher intensities than $\tau_\text{raw}$.
Next, we motion compensate the extracted point cloud $\mathcal{P}^t$ in 3D with respect to the earliest point in the point cloud, using orientation data obtained through spherical linear interpolation.
The deskewed point cloud $\mathcal{\hat{P}}^t$ is then processed through a tilt-based filter, where we first compute the vertical distance $\Delta d\left(\mathbf{p}_i,q_r\right)$ between each individual point $\mathbf{p}_i\in\mathcal{\hat{P}}^t$ and a rotation axis defined by the pitch and roll from the relative quaternion $q_r$, where $i$ is a point's index.
We then weight each points with the Cauchy function
\begin{equation}
    w_i=\frac{1}{1+\left(\frac{\Delta d\left(\mathbf{p}_i,q_r\right)}{\gamma}\right)^2},
\end{equation}
where $w_i$ are per-point weights and $\gamma$ is the scale factor.
We apply a hard threshold $\tau_\text{tilt}$ removing all points too far from the axis, producing $\mathcal{\tilde{P}}_t$.
This filtered point cloud is then matched to the reference submap $\mathcal{Q}$ with Point-to-point \ac{ICP} registration.
Both point clouds are subsampled with a spatial filter with voxel size $d_\text{voxel}$, where we compute and store the centroid in each voxel, before a $k_\text{nn}$ nearest neighbor correspondence search.
We use a constant velocity model, together with the estimated yaw angle $\psi$, as a prior for the next submap lookup and the point-to-point registration.
The deskewed point cloud $\mathcal{\hat{P}}_t$ and the corrected 2D~pose $\mathbf{p}^t$ are fed into the submap management system, which either merges $\mathcal{\hat{P}}_t$ into an existing submap, or starts a new one based on a distance or a tilt difference, utilizing the previously introduced distance $r_\text{submap}$ and roll and pitch $\tau_\text{tilt}$ thresholds.

\section{Experiments}
\begin{figure}
    \centering
    \includegraphics[width=\linewidth]{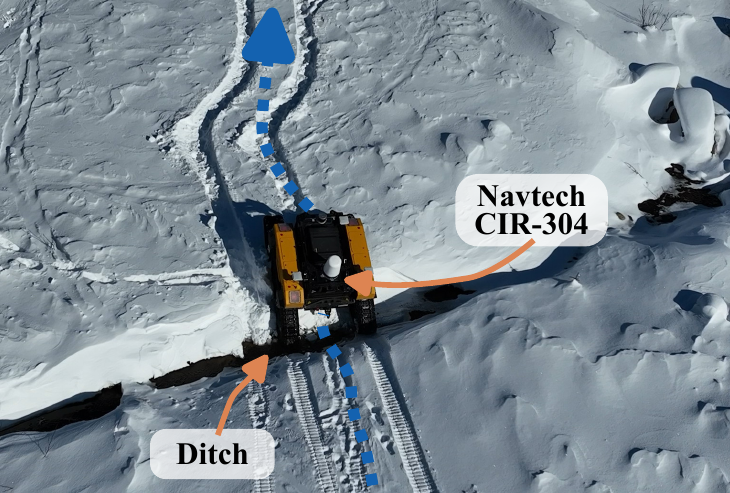}
    \caption{The \acl{UGV} equipped with a Navtech CIR-304 radar and a Vectornav VN-100 IMU, passing through a ditch, causing a sudden change in the vehicle's orientation.
    }
    \label{fig:ditch}
\end{figure}

We evaluate our method, together with three state-of-the-art approaches (CFEAR~\cite{Adolfsson2023_cfear}, ORORA~\cite{Lim2023_orora} and \ac{RTR}~\cite{Qiao2025}), on two sequences from the \ac{FoMo} dataset \cite{Boxan2026_fomo}.
The \ac{UGV} used to record the data was a Clearpath Warthog, equipped with a Navtech CIR-304 radar and a VectorNav VN-100 \ac{IMU}.
The performance of the evaluated algorithms is assessed on two distinct trajectories: the \texttt{Red} and \texttt{Orange} loops.
The \texttt{Red} loop is a \qty{300}{\meter} circuit around a building on flat, paved terrain, serving as a controlled baseline for urban-like conditions.
In contrast, the \texttt{Orange} loop is a \qty{2}{\kilo\meter} trajectory characterized by high-complexity terrain.
This sequence includes a stone quarry, depicted in \autoref{fig:ditch}, with significant pitch and roll variations, as well as long forest corridors where geometry is sparse and features are limited.
Additionally, we evaluate the methods on sequences from four seasons, therefore testing their robustness against environmental variations, such as snow accumulation of up to \qty{1}{\meter}.
The total number of tested sequences is 11 for the \texttt{Red} trajectory and 9 for the \texttt{Orange} trajectory.
The proposed method was tuned using a minimal training set: a single \texttt{Red} trajectory recorded in August and a \qty{200}{\meter} segment of the stone quarry sequence from January. The resulting parameters are detailed in \autoref{tab:params}.
Our implementation runs in ROS2 and relies on libpointmatcher \cite{Pomerleau2012_comp} for point cloud registration.

\begin{table}[ht]
\centering
\caption{Radar-Odometry Pipeline Parameters}
\label{tab:params}
\begin{tabularx}{\columnwidth}{llX}
\toprule
\textbf{Parameter} & \textbf{Value} & \textbf{Description} \\
\midrule
$k$                  & $10$    & High-intensity peaks per azimuth. \\
$r_{\min}$           & $5.0$   & Min range to filter near-field noise. \\
$r_{\max}$           & $100.0$ & Max operational sensing range. \\
$\tau_{\text{raw}}$  & $60$    & Raw intensity return threshold. \\
$d_{\text{voxel}}$   & $1.0$   & Voxel size for spatial downsampling. \\
$\theta_{\text{tilt}}$ & $3.0$   & Orientation threshold for tilt detection. \\
$\gamma$             & $3.5$   & Cauchy cost function scale parameter. \\
$r_{\text{submap}}$  & $20.0$  & Distance for submap search and updates. \\
$\tau_{\text{tilt}}$ & $0.8$   & Weight threshold for tilt-based filtering. \\
$k_{\text{nn}}$      & $4$     & Neighbors for ICP local search. \\
\bottomrule
\end{tabularx}
\end{table}

\section{Results}
\label{sec:results}

\begin{figure}[htb]
    \centering
    \includegraphics[width=\linewidth]{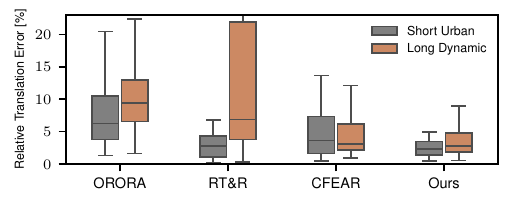}
    \caption{Comparison of the performance of the four tested methods on two experiments:
    a short urban loop of \qty{300}{\meter} with distinct features, and a \qty{2}{\kilo\meter}-long dynamic loop with varying vehicle orientation and tougher terrain.}
    \label{fig:boxplot}
\end{figure}

In this section, we present the results obtained on the total of 20~runs from the \ac{FoMo} dataset.
The aggregated \ac{RTE} is presented in \autoref{fig:boxplot}.
The figure shows the distribution of \ac{RTE} computed on \qty{100}{\meter}-long segments for the two trajectory types, the short urban circuit \texttt{Red} in gray and the long dynamic loop \texttt{Orange} in orange.
With the exception of CFEAR, all methods exhibited degraded performance on the dynamic loop compared to the urban loop.
We attribute this discrepancy to a combination of the evaluation metric and the geometric properties of each trajectory.
The compact \text{Red} loop features turns within each \qty{100}{\meter} window, amplifying the impact of heading estimation errors, whereas the long \texttt{Orange} loop's extended straight segments may underweight rotational inaccuracies.
Looking at the results, while ORORA suffers from a wrong heading estimation in both experiments, \ac{RTR} relies on an orientation prior from the angular velocity around the z-axis of the \ac{IMU}, $\omega_z$.
Although this approach gives good results in the case of the flat urban circuit, the error grows as the single axis is not sufficient to estimate the \ac{UGV}'s heading in the highly dynamic loop.
Our method, on the other hand, uses both $\mathbf{\omega}$ and $\mathbf{a}$, constraining the heading error in both experiments.
Additionally, we notice that \ac{RTR} fails in 2/9 runs of the dynamic trajectory due to high tilt difference between consecutive scans.
While CFEAR slightly improves from \qty{3.6}{\percent} to \qty{3.1}{\percent} median error between experiments, our method degrades from \qty{2.3}{\percent} to \qty{2.8}{\percent}, still making it the best overall performing method.

\begin{figure}[htb]
    \centering
    \includegraphics[width=\linewidth]{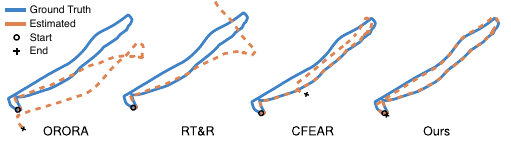}
    \caption{Output trajectory of the four evaluated methods, together with the \acl{GT}, originating from the Long dynamic loop in March.
    Note that \ac{RTR} is cropped on the top.}
    \label{fig:orange}
\end{figure}

In \autoref{fig:orange}, we provide an example of a single run on the long dynamic loop \texttt{Orange} in March.
This run, taking place in the later part of winter, saw the highest snow accumulation of all data in the tested sequences.
Consequently, the methods had to deal with high tilt dynamic, lateral slip and a corridor effect caused by tall snowbanks around forest roads.
Both ORORA and \ac{RTR} failed to correctly estimate the \ac{UGV}'s orientation, with ORORA crossing the original path and \ac{RTR} diverging from the \ac{GT} in a sharp angle.
CFEAR, on the other hand, suffered from the corridor effect caused by snowbanks on a plowed section of the road.
Our method completed the loop, although still finishing \qty{23}{\meter} from the \acl{GT} position.

\begin{figure}[htb]
    \centering
    \includegraphics[width=\linewidth]{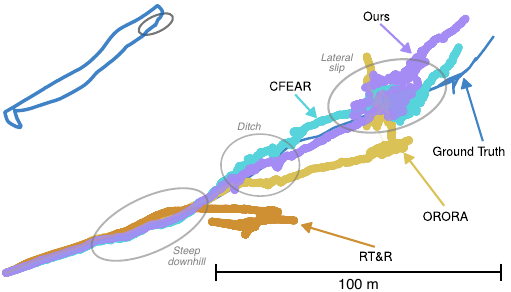}
    \caption{Detailed view on the estimated trajectory from the four evaluated methods, together with the \acl{GT}, on a dynamic section in a stone quarry from the Long dynamic loop in March.}
    \label{fig:quarry}
\end{figure}

While the previous results might indicate that rotating \ac{FMCW} radar odometry can be sufficient for tasks that do not require absolute positioning, such as \ac{TaR}, \autoref{fig:quarry} provides a different picture.
The figure shows a zoomed-in area from the long dynamic loop, covering approximately \qty{200}{\meter} in a stone quarry.
The sequence, taking place in March, starts at a highest point of the trajectory, before going steeply downhill.
Next, the vehicle passes through the ditch displayed in \autoref{fig:ditch}.
Finally, the \ac{UGV} climbs uphill, experiencing high lateral slip.
\ac{RTR} failed to correctly estimate the vehicle's heading already in the downhill section.
ORORA diverged from the \ac{GT} in the ditch section, and again in the high slip area.
Compared to CFEAR, our method saw smaller deviation in the ditch section and recovered at a more accurate orientation after the high slip area.
Nevertheless, all evaluated methods exhibit high-frequency noise and discontinuous \textit{jumps} in the estimated pose.
Such artifacts would either produce reference trajectories that violate the vehicle’s kinematic constraints or necessitate aggressive smoothing.
However, applying such filters risks degrading the trajectory's fidelity in other segments—for instance, by inadvertently \textit{cutting} corners during sharp maneuvers.

\section{Conclusion and Future work}
In this work, we presented a novel radar-inertial odometry method specifically tailored for high-complexity trajectories characterized by significant pitch and roll variations.
We benchmarked our approach against three state-of-the-art radar odometry frameworks across two experimental scenarios: a controlled urban loop and a long-duration dynamic trajectory.
Our results demonstrate that the proposed method achieves state-of-the-art performance in urban environments while outperforming existing baselines in dynamic subarctic conditions.
Furthermore, we discussed the individual method's behavior in a highly dynamic \qty{200}{\meter}-long section that includes steep downhill, a ditch formed by a small creek and high lateral slip area.

Moving forward, we intend to further enhance our pipeline by incorporating translational priors into the motion-compensation module.
We will also conduct an extensive ablation study to isolate the individual contributions of our tilt-based filtering and submap-search strategies.
To broaden the scope of our evaluation, we plan to integrate a dense radar odometry method into our benchmarking suite and validate the robustness of our framework on standard, large-scale urban driving datasets.

\addtolength{\textheight}{0cm}   

\section*{ACKNOWLEDGMENT}


\printbibliography

\end{document}

%% file: preamble.tex
\usepackage[
    backend=biber,
    style=ieee,
    sorting=none,
    natbib=true,
    doi=false,
    isbn=false,
    url=false,
    eprint=false,
    maxcitenames=1,
    mincitenames=1
]{biblatex}

\usepackage{amssymb,amsfonts,amsmath,amscd}

\usepackage{bm}

\usepackage[colorlinks,bookmarksopen,bookmarksnumbered,citecolor=red,urlcolor=red]{hyperref}

\renewcommand*{\thesubsection}{\thesection.\Alph{subsection}}
\usepackage[english,noabbrev,nameinlink]{cleveref}

\usepackage[printonlyused]{acronym}

\usepackage{siunitx}
\usepackage[all]{nowidow}

\usepackage[dvipsnames]{xcolor}

\usepackage{lipsum}

\usepackage{xspace} 

\usepackage{xurl}

\usepackage{epstopdf}

\usepackage{import}

\usepackage{graphicx}

\usepackage{booktabs}

\usepackage{tabularx}
\usepackage{multirow, multicol}

\usepackage{algorithm}
\usepackage{algpseudocode}

\usepackage{dirtree}

\usepackage[english]{datetime2}
\DTMsetdatestyle{ddmmyyyy}

\newcommand{\bbm}{\begin{bmatrix}}
\newcommand{\ebm}{\end{bmatrix}}

\acrodef{ICP}{Iterative Closest Point} 
\acrodef{MOCAP}{Motion Capture} 
\acrodef{SLAM}{Simultaneous Localization and Mapping}
\acrodef{VSLAM}{Visual SLAM} 
\acrodef{IMU}{Inertial Measurement Unit} 
\acrodef{GT}{Ground Truth} 
\acrodef{FoMo}{For\^{e}t Montmorency} 
\acrodef{SDK}{Software Development Kit}
\acrodef{GNSS}{Global Navigation Satellite System} 
\acrodef{UGV}{Uncrewed Ground Vehicle} 
\acrodef{UAV}{Uncrewed Aerial Vehicle} 
\acrodef{FMCW}{Frequency Modulated Continuous Wave}
\acrodef{PTP}[PTP]{IEEE1588 Precision Time Protocol}
\acrodef{NMEA}[NMEA]{National Marine Electronics Association}
\acrodef{PPK}[PPK]{Post Processed Kinematic}
\acrodef{RTK}[RTK]{Real Time Kinematic}
\acrodef{PPS}[PPS]{pulse-per-second}
\acrodef{DL}[DL]{Deep Learning}
\acrodef{GMSL2}[GMSL2]{Gigabit Multimedia Serial Link 2}
\acrodef{INS}{Inertial Navigation System}
\acrodef{AE}{Auto Exposure}
\acrodef{ROI}{Rectangle of Interest}
\acrodef{RANSAC}{Random Sample Consensus}
\acrodef{RTS}{Robotics Total Station}
\acrodef{BEV}{Bird's-eye view}
\acrodef{ETH}{Ethernet}
\acrodef{RINEX}{Receiver Independent Exchange Format}
\acrodef{CORS}{Continuously Operating Reference Station}
\acrodef{DoF}{Degree of Freedom}
\acrodefplural{DoF}[DoFs]{Degrees of Freedom}
\acrodef{RTR}[RT\&R]{Radar Teach and Repeat}
\acrodef{LTR}[LT\&R]{Lidar Teach and Repeat}
\acrodef{RGTR}[RT\&R]{Radar-Gyro Teach and Repeat}
\acrodef{TaR}{Teach and Repeat}
\acrodef{NLP}{Natural Language Processing}
\acrodef{FOV}{Field of View}
\acrodef{RPE}{Relative Pose Error}
\acrodef{IDD}{Ideal Differential Drive}
\acrodef{APE}{Absolute Pose Error}
\acrodef{ATE}{Absolute Trajectory Error}
\acrodef{RTE}{Relative Translation Error}
\acrodef{RE}{Relative Error}
\acrodef{PGO}{Pose Graph Optimization}
\acrodef{BA}{Bundle Adjustment}
\acrodef{SURF}{Speeded-up Robust Features}
\acrodef{IQR}{Inter-quartile Range}
\acrodef{SARTE}{Sequentially Aligned Relative Trajectory Error}
\acrodef{RTDE}{Relative Traveled Distance Error}
\acrodef{LR}{Localization Ratio}
\acrodef{ILR}{Inverse Localization Ratio}
\acrodef{CFAR}{Constant False Alarm Ratio}

\usepackage{fontawesome5}
\usepackage{xcolor}

\makeatletter
\newcommand{\showfont}{%
  Font family: \f@family, %
  Font series: \f@series, %
  Font shape: \f@shape, %
  Font size: \f@size pt%
}
\makeatother

\DeclareSIUnit{\degree}{°}

\usepackage[subpreambles=false]{standalone}

\usepackage{tikz}
\usepackage{amsmath}
\usepackage{amssymb}
\usetikzlibrary{arrows.meta, positioning, calc}

\usepackage{tikz}
\usepackage{amsmath}
\usepackage{amssymb}
\usetikzlibrary{arrows.meta, positioning, calc}

\definecolor{polarScanBlue}{HTML}{ 61CBF4}
\definecolor{featuresScanBlue}{HTML}{ 4E95D9}
\definecolor{filteredScanBlue}{HTML}{ 5175EA}

\definecolor{col-imu}{RGB}{56, 102, 65}
\definecolor{col-radar}{RGB}{70, 110, 160}
\definecolor{col-submap}{RGB}{210, 120, 50}
\definecolor{col-reg}{RGB}{90, 90, 95}

\tikzset{
  base/.style={
    rectangle, rounded corners=5pt, draw=none,
    text=white, font=\sffamily\small,
    minimum height=1.15cm, align=center,
    inner xsep=8pt, inner ysep=6pt
  },
  imu/.style   = {base, fill=col-imu},
  radar/.style = {base, fill=col-radar},
  submap/.style= {base, fill=col-submap},
  reg/.style   = {base, fill=col-reg},
  arr/.style={-{Stealth[length=6pt,width=5pt]}, very thick, black},
  darr/.style={arr, dashed},
  lbl/.style={font=\sffamily\small, midway}
}


%% file: diagram.tex
\begin{tikzpicture}[
  node distance = 0.55cm and 0.52cm
]

  \node[imu,  minimum width=2.7cm, text width=2.5cm] (imu)
        {%
        Raw IMU data\\
        200\,Hz};

  \node[imu,  minimum width=2.2cm, right=1.4cm of imu] (madg)
        {Madgwick filter};

  \node[radar, minimum width=2.7cm, below=0.7cm of imu, text width=2.5cm] (radar)
        {Polar radar scan\\4\,Hz};

  \node[radar, minimum width=2.4cm, right=1.4cm of radar] (feat)
        {Feature extraction\\$k$-strongest};

  \node[radar, minimum width=2.0cm, right=0.6cm of feat]  (desk)
        {3D deskewing};

  \node[radar, minimum width=2.3cm, right=0.6cm of desk]  (tilt)
        {Tilt-based\\filtering};

  \node[reg, minimum width=2.5cm, right=0.6cm of tilt]  (p2p)
        {Point-to-point\\registration};

  \node[submap, minimum width=2.2cm, right=0.6cm of p2p] (mgmt)
        {Submap\\management};

  \node[submap, minimum width=2.2cm, above=0.7cm of tilt] (lookup)
        {Submap\\lookup};
        
  \node[reg,   minimum width=2.4cm, right=2.3cm of lookup] (prior)
        {Prior\\estimation};

  \draw[arr, polarScanBlue] (radar) -- node[lbl,black, above]{$Z_{N_a \times N_r}$} (feat);
  \draw[arr, featuresScanBlue] (feat)  -- node[lbl,black,above]{$\mathcal{P}^t$}               (desk);
  \draw[arr] (desk)  -- node[lbl,above]{$\hat{\mathcal{P}}^t$}         (tilt);
  \draw[arr, filteredScanBlue] (tilt)  -- node[lbl,black,above]{$\tilde{\mathcal{P}}^t$}       (p2p);

  \draw[arr] (p2p.north east) ++(0.,-0.3) -| 
      node[lbl, above, xshift=-5pt, yshift=5pt] {$\mathbf{\dot{p}}^t$} 
      (prior);
  
  \draw[arr] ($(p2p.east) + (0,-0.3)$) -- node[lbl,above]{$\mathbf{p}^t$} ($(mgmt.west) + (0,-0.3)$);
  \draw[arr] (imu) -- node[lbl,above]{$\boldsymbol{a},\,\boldsymbol{\omega}$} (madg);

  \draw[arr] (madg.east) -- node[lbl,above]{$q \in \mathbb{H}$} (lookup.west);

 \draw[darr] (lookup.north) 
    -- node[above, pos=0.5, font=\sffamily\small, xshift=15pt, yshift=3pt]{Yaw angle $\psi_r$} 
       ($(lookup.north) + (0.0, 0.4)$)        
    -| (prior.north);                        


  \draw[arr] (madg.east) -| (desk.north);

  \draw[arr] (lookup.south) -- node[lbl,left]{$q_r\in\mathbb{H}$} (tilt.north);

  \draw[arr] (lookup.south east) ++(-0.3,0) -- ++(0,-0.3) -| node[lbl, pos=0.5, right]{$\mathcal{Q}$} (p2p.north);

  \draw[arr] (prior.west) -- (lookup.east);
  
  \draw[arr] (desk.east) -- ++(0.3,0) -- ++(0.0,-0.9) -| (mgmt.south);

\end{tikzpicture}